\def\year{2021}\relax
\documentclass[letterpaper]{article} % DO NOT CHANGE THIS
\usepackage{aaai21}  % DO NOT CHANGE THIS
\usepackage{times}  % DO NOT CHANGE THIS
\usepackage{helvet} % DO NOT CHANGE THIS
\usepackage{courier}  % DO NOT CHANGE THIS
\usepackage[hyphens]{url}  % DO NOT CHANGE THIS
\usepackage{graphicx} % DO NOT CHANGE THIS
\usepackage{natbib}  % DO NOT CHANGE THIS AND DO NOT ADD ANY OPTIONS TO IT
\usepackage{caption} % DO NOT CHANGE THIS AND DO NOT ADD ANY OPTIONS TO IT
\usepackage{epsfig}
\usepackage{amsmath}
\usepackage{amssymb}
\usepackage{algorithm}
\usepackage{algorithmic}
\usepackage{enumerate}
\usepackage{multirow}
\usepackage{array}
\usepackage{cprotect}
\usepackage{mathrsfs}
\usepackage{subfigure}
\usepackage{threeparttable}
\usepackage{fancyvrb}

\title{Stratified Rule-Aware Network for Abstract Visual Reasoning}

\author {
        Sheng Hu,\textsuperscript{\rm 1,}\footnote{Equal contribution}
        Yuqing Ma,\textsuperscript{\rm 1,}\footnotemark[\value{footnote}]
        Xianglong Liu,\textsuperscript{\rm 1,2,}\footnote{Corresponding author}   
        Yanlu Wei,\textsuperscript{\rm 1}
        Shihao Bai\textsuperscript{\rm 1} \\
}
\affiliations {
    \textsuperscript{\rm 1} State Key Laboratory of Software Development Environment, Beihang University, Beijing, China \\
    \textsuperscript{\rm 2} Beijing Advanced Innovation Center for Big Data-Based Precision Medicine, Beihang University, Beijing, China \\
    husheng\_7@163.com, \{mayuqing,xlliu\}@nlsde.buaa.edu.cn, \{weiyanlu,16061167\}@buaa.edu.cn
}

\begin{document}

\maketitle

\begin{abstract}
Abstract reasoning refers to the ability to analyze information, discover rules at an intangible level, and solve problems in innovative ways. Raven's Progressive Matrices (RPM) test is typically used to examine the capability of abstract reasoning. The subject is asked to identify the correct choice from the answer set to fill the missing panel at the bottom right of RPM (e.g., a 3$\times$3 matrix), following the underlying rules inside the matrix. Recent studies, taking advantage of Convolutional Neural Networks (CNNs), have achieved encouraging progress to accomplish the RPM test. However, they partly ignore necessary inductive biases of RPM solver, such as order sensitivity within each row/column and incremental rule induction. To address this problem, in this paper we propose a Stratified Rule-Aware Network (SRAN) to generate the rule embeddings for two input sequences. Our SRAN learns multiple granularity rule embeddings at different levels, and incrementally integrates the stratified embedding flows through a gated fusion module. With the help of embeddings, a rule similarity metric is applied to guarantee that SRAN can not only be trained using a tuplet loss but also infer the best answer efficiently. We further point out the severe defects existing in the popular RAVEN dataset for RPM test, which prevent from the fair evaluation of the abstract reasoning ability. To fix the defects, we propose an answer set generation algorithm called Attribute Bisection Tree (ABT), forming an improved dataset named Impartial-RAVEN (I-RAVEN for short). Extensive experiments are conducted on both PGM and I-RAVEN datasets, showing that our SRAN outperforms the state-of-the-art models by a considerable margin.
\end{abstract}

\section{Introduction}

Abstract reasoning, also known as inductive reasoning, refers to the ability to analyze information, discover rules at an intangible level, and solve problems in innovative ways. This type of reasoning, as the foundation for human intelligence, helps human understand the world. 
It has been generally regarded and pursued as a critical component to the development of artificial intelligence during the past decades, and has attracted increasing attention in recent years.
Raven's Progressive Matrices (RPM) test~\cite{Raven1938,Carpenter1990What,raven2000raven,kunda2013computational,strannegaard2013anthropomorphic} is one of the highly accepted and well-studied tools to examine the ability of abstract reasoning, which is believed as a good estimate of the real intelligence~\cite{Carpenter1990What}.
\begin{figure}[tp!]
	\center
	\includegraphics[width=1.0\linewidth]{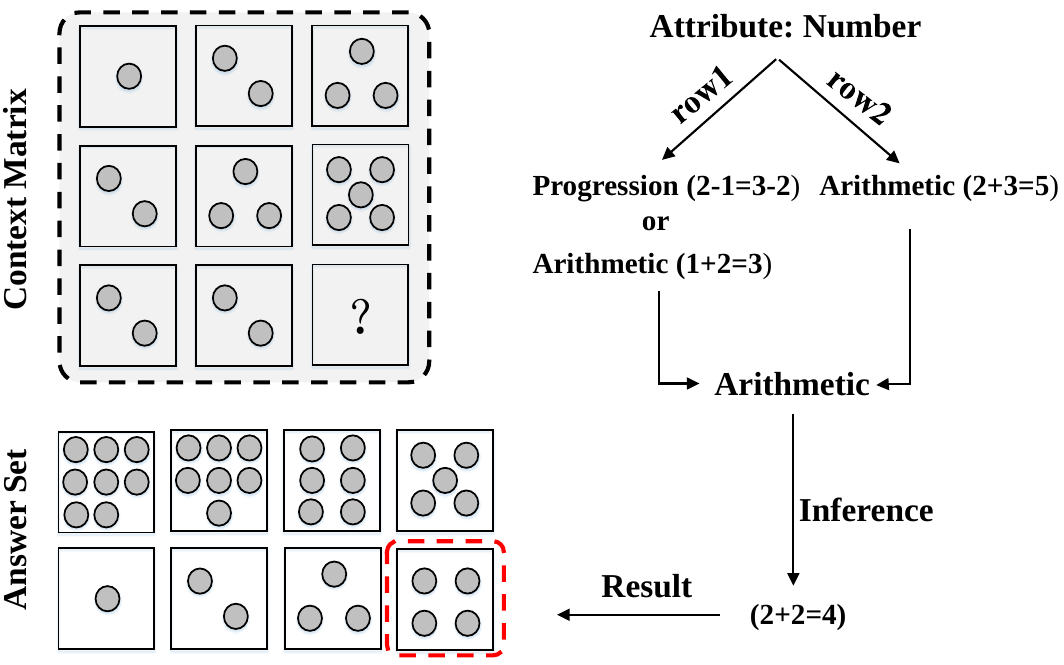}
	\caption{An example of RPM question and the human strategy to solve it. The underlying rule on the number of circles could be \textit{Progression} (2-1=3-2) or \textit{Arithmetic} (1+2=3) along row 1, and \textit{Arithmetic} (2+3=5) along row 2. Therefore the dominant rule is \textit{Arithmetic}. Apply it to the third row to figure out the answer (2+2=4). Besides, no viable rule can be found along the columns}
	\label{human-stra}
	%\vspace{-0.1in}
\end{figure}
An illustration of RPM is shown in Figure~\ref{human-stra}, where usually the test-taker is presented with a 3$\times$3 matrix with the bottom right panel left blank.
The goal is to choose one image from an answer set of eight candidates to complete the matrix correctly, namely satisfying the underlying rules in the matrix. Subjects accomplish this by looking into the first two rows/columns and inducing the dominant rules which govern the attributes in those panels. The obtained rules can then be applied to the last row/column to figure out which answer belongs to the blank panel.

Computational models for RPM in the cognitive science community access symbolic representations of the images~\cite{Carpenter1990What,LovettModeling,lovett2010structure,Lovett2010Solving}. Recently there has been some success with end-to-end learning methods trying to accomplish abstract reasoning on RPM test~\cite{hoshen2017iq,BarrettMeasuring,steenbrugge2018improving,ZhangRAVEN,zhanglearning,zheng2019abstract,van2019disentangled,wang2020abstract}, inspired by the progress of computer vision tasks~\cite{krizhevsky2012imagenet,simonyan2014very,szegedy2015going,he2016deep} and boosted by the large-scale PGM~\cite{BarrettMeasuring} and RAVEN~\cite{ZhangRAVEN} datasets. 
Typical works including CoPINet~\cite{ZhangRAVEN}, LEN~\cite{zheng2019abstract}, and MXGNet~\cite{wang2020abstract} followed the paradigm that predicts a classification score for each multiple-choice panel based on the relations inside each row/column, showing great potential to solve RPM test.
However, these models partly ignore the important characteristics for RPM, such as the permutation invariance~\cite{zhanglearning}, the order sensitivity of panels inside a row/column, etc.
Previous work~\cite{wang2020abstract} specially mentions that they do not choose a permutation-invariant structure because it leads to severe ‘overfitting’ on the RAVEN dataset. We will discuss this phenomenon in later sections.
What is even worse, directly extracting the relations, without considering the incremental rule induction mechanism widely adopted in human cognitive systems~\cite{Carpenter1990What}, inevitably leads to inferior performance. 

To achieve reliable and efficient abstract reasoning, in this paper we develop a powerful architecture called Stratified Rule-Aware Network (SRAN) that naturally integrates the indispensable inductive biases, including order sensitivity, permutation invariance, and incremental rule induction. 
SRAN takes two rows/columns as input and learns stratified rule embeddings at different levels, namely cell-wise, individual-wise, and ecological hierarchy.
These multiple granularity embeddings are incrementally integrated via a gate fusion module, which naturally preserves the order sensitivity of panels and maps the inputs to a rule embedding space. 
With the help of the embeddings, we further introduce a rule similarity metric, based on which SRAN can not only be well trained using a tuplet loss but also infer the best answer efficiently. This framework resembles the human strategy for RPM shown in Figure \ref{human-stra}. 

To fairly evaluate the abstract reasoning ability, we also design a general algorithm named Attribute Bisection Tree (ABT) to generate an impartial answer set for any attribute-based RPM question. We point out and further fix the underlying defects of the commonly-used RAVEN~\cite{ZhangRAVEN} dataset, where the correct answer could be inferred even without the presence of the context matrix. Therefore, we introduce an improved dataset named Impartial-RAVEN (I-RAVEN) to fairly evaluate the abstract reasoning capability of RPM solvers. 

To the best of our knowledge, the proposed SRAN is the first RPM solver to induce rule embeddings which are discriminative and measurable.
We are also the first to point out and fix the defects of the misleading benchmark RAVEN, and generate an impartial dataset I-RAVEN based on our ABT algorithm. 
Extensive experiments conducted on widely used dataset PGM and our improved I-RAVEN show that SRAN outperforms state-of-the-art methods by a considerable margin, e.g. 60.8\% accuracy compared to the second best 46.1\% on I-RAVEN.

\section{Our Approach}
In this section, we first give a formal definition of the abstract reasoning task on the RPM test. Then we introduce the inductive-biased framework, and present our Stratified Rule-Aware Network (SRAN). Finally, we demonstrate the learning and inference process of the proposed model.

\subsection{Preliminary}\label{sec:preliminary}
For a common RPM question, usually a 3$\times$3 matrix $\mathbf{M}^{-}$ is given, with bottom right context panel left blank. $\Omega$ denotes the answer set with $N$ multiple-choice panels, where typically $N$=8. The dominant rules governing the features inside the matrix could be induced from the first two intact rows/columns. The goal is to select a multiple-choice panel $\omega \in \Omega$ to complete the context matrix $\mathbf{M}^{-}$, maintaining the dominant rule inside of the context matrix. 

We define the completed matrix with a multiple-choice panel $\omega$ infilled as $\mathbf{M}$, where $\mathbf{M}_i$ is denoted as the $i$-th row, and $\mathbf{m}_{ij}$ indicates the panel in $i$-th row and $j$-th column. Intuitively, $\mathbf{M}$ is almost the same as $\mathbf{M}^{-}$, except for $\mathbf{m}_{33}=\omega$ while the corresponding element missing in $\mathbf{M}^{-}$. In fact, whether rules exist in rows or columns is uncertain. 
Therefore, our framework induces both the row-wise rule representation and the column-wise representation in the same way. In order to simplify the notation, we only take the induction of the row-wise rule representation as example. 

\subsection{The Reasoning Framework}\label{sec:framework}
Based on the necessary inductive biases for RPM, we develop a novel abstract reasoning architecture named Stratified Rule-Aware Network (SRAN).
Given two input rows $\mathbf{M}_i, \mathbf{M}_j$, the proposed framework embeds the input into multiple granularity embeddings using a stratified rule embedding module $\mathbb{E}$. 
Named after biological organizations~\cite{parent1996living}, $\mathbb{E}$ consists of three hierarchies including cell-wise network $\mathbb{E}_{\text{cell}}$, individual-wise network $\mathbb{E}_{\text{ind}}$, and ecological network $\mathbb{E}_{\text{eco}}$. With the multiple granularity rule embeddings, the gated embedding fusion module $\mathbb{G}$ will incrementally integrate these stratified embedding flows and map the two input sequences $\mathbf{M}_i$ and $\mathbf{M}_j$ to a discriminative rule embedding $\mathbf{r}^{(3)}_{ij}$, while maintaining the order sensitivity and permutation invariance.
We further introduce a rule similarity metric $\mathcal{D}$ to estimate the similarity between the rule representations.
The correct answer can be predicted by choosing the multiple-choice panel within the shortest distance to the dominant rule generated by the first two rows in the matrix.

\subsection{Stratified Rule Embedding}\label{rule_embed}
As we all know, organization of behaviour into a nested hierarchy of tasks is characteristic of purposive cognition in humans. The prevalent Convolution Neural Network inspired by the human visual system, is a stratified model itself, with the projection from each layer showing the hierarchical nature of features. The bottom layers extract low-level features, such as texture, edge, etc, while the top layers abstract high-level semantic information from the low-level information transmitted from the bottom layers. 

However, without specifying information from different levels, it is hard for CNN to figure out different hierarchies, and thus fail to obtain robust and discriminative representations.
Therefore, it would be better to feed the input of different hierarchies explicitly and extract rule representations from different granularity with artificial guidance. Motivated by that, we deploy a stratified rule embedding module, consisting of cell-wise hierarchy, individual-wise hierarchy, and ecological hierarchy.

\subsubsection{Cell-wise Hierarchy}
The network of the cell-wise hierarchy $\mathbb{E}_{\text{cell}}$ takes each panel as input and recognize the attributes of inside graphical elements. It handles each panel independently without considering the difference or correlations among panels inside the matrix. Therefore, it observes the information from the most detailed perspective. We obtain the cell-wise rule representation for each input panel:
\begin{equation}
	\mathbf{x}_{ij}=\mathbb{E}_{\text{cell}}(\mathbf{m}_{ij}).
\end{equation}

\subsubsection{Individual-wise Hierarchy}
Moreover, the network of individual hierarchy takes each row as input. It begins to take the correlations among panels of the same row into consideration, and encode the entire row with a compact embedding, rather than simply combining each panel. In this way, the rule embedding process for each panel is coupled and interacts with each other. Intuitively, each row may contain multiple plausible rules. In this hierarchy, the framework extracts intermediate rule embedding for each row individually, which still ignores the comprehensive information from the matrix perspective, especially the correlations across rows. The individual-wise rule embedding $\mathbf{y}_{i}$ is denoted as:
\begin{equation}
	\mathbf{y}_{i}=\mathbb{E}_{\text{ind}}(\mathbf{M}_i). 
\end{equation}

\subsubsection{Ecological Hierarchy}
Furthermore, the network of the ecological hierarchy takes the two rows together as input and jointly learns the rule patterns underlying the two rows. As we mentioned before, in the individual hierarchy, the framework extracts intermediate rule embedding for each row, without considering the interaction between two rows. The rule that exists in one row may not lie in another. Therefore, to obtain the shared rule patterns between the two rows, it is essential to put these two rows together and jointly learn the features from an ecological level. Thus the shared rule embedding is obtained as follows:
\begin{equation}
	\mathbf{z}_{ij}=\mathbb{E}_{\text{eco}}([\mathbf{M}_i, \mathbf{M}_j]),
\end{equation}
where $[\cdot, \cdot]$ denotes the concatenating operation.

\subsection{Gated Embedding Fusion}\label{gated}
Since the rule embeddings at different levels focus on different attributes or patterns, to generate one discriminative representation for the rule, we should aggregate the multiple granularity embeddings. Due to the requirement that the aggregation should preserve the order of cell-wise rule embeddings and be permutation-invariant to the individual-wise ones, we propose a stratified rule embedding learning method named gated embedding fusion module, which is responsible for gradually aggregating the multiple granularity embeddings.

Specifically, we define a gate function $\varphi$ to fuse the rule embeddings from different hierarchies. It concatenates all the inputs and encodes into a single embedding using fully connected layers. The gate function is similar to the attention mechanism, which detects and concentrates on the useful features according to the task. Even for the same attribute, they may focus on different facets. Based on the gate function, our gated embedding fusion module could regulate the flow of rule embeddings into the framework and make the utmost of their complementary information.

\begin{figure}[tp!]
	\centering
	\includegraphics[width=1.0\linewidth]{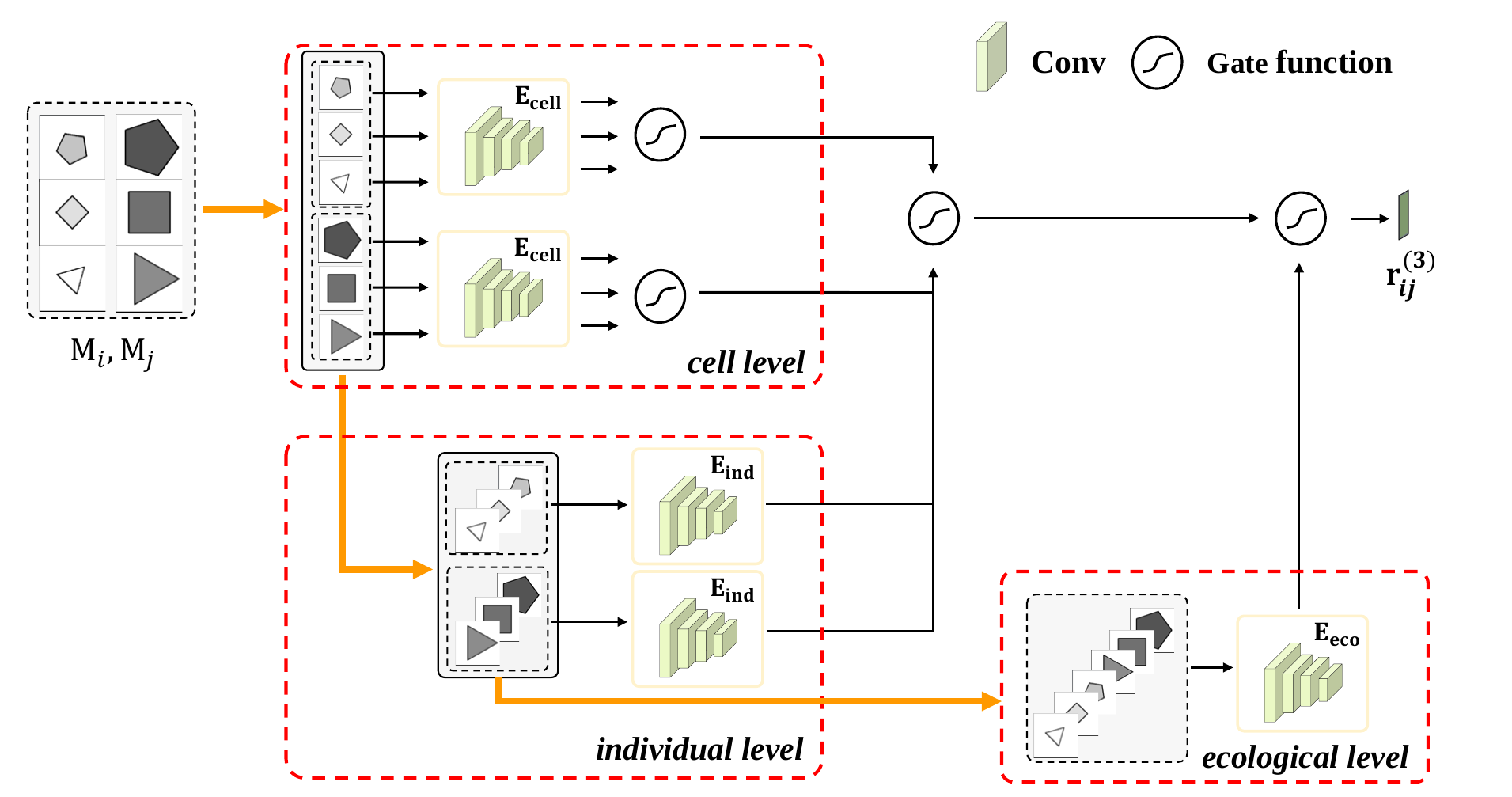} 
	\caption{The architecture of SRAN, consisting of a hierarchical rule embedding module and a gated embedding fusion module. Given two row sequences as input, it outputs the rule embedding}
	\label{network-SRAN}
	%\vspace{-0.20in}
\end{figure}

At the cell level, after obtaining cell-wise rule embeddings for panels in $i$-th row $\mathbf{M}_i$, the module aggregates them to infer a row-wise rule embedding $\mathbf{r}_{i}^{(1)}$:
\begin{equation}
	\mathbf{r}_{i}^{(1)}=\varphi_1(\mathbf{x}_{i1}, \mathbf{x}_{i2}, \mathbf{x}_{i3}),
\end{equation}
Similarly, we obtain $\mathbf{r}_{j}^{(1)}$ for the $j$-th row $\mathbf{M}_j$. The fused embedding integrates different types of attributes in the panels.

At the individual level, intuitively both $\mathbf{r}^{(1)}_{i}$ and $\mathbf{y}_{i}$ are the row-wise embeddings corresponding to the $i$-th row, but convey the different granularity rule information. We further fuse them, and jointly mine the shared rules contained in the $i$-th and $j$-th row :
\begin{equation}
	\mathbf{r}^{(2)}_{ij} = \varphi_2(\mathbf{r}_{i}^{(1)}, \mathbf{y}_{i}, \mathbf{r}_{j}^{(1)},  \mathbf{y}_{j}).
\end{equation}

At the ecological level, similarly we can further combine fused embedding $\mathbf{r}^{(2)}_{ij}$ and $\mathbf{z}_{ij}$ using the gate fusion function, abstracting the final rule embedding:
\begin{equation}
	\mathbf{r}_{ij}^{(3)}=\varphi_3(\mathbf{r}^{(2)}_{ij}, \mathbf{z}_{ij}).
\end{equation}

To make sure the framework is permutation-invariant to the input rows, we exchange the concatenation order of the two input rows and average the output rule embeddings. 
This invariance ensures that, the rule embedding respects the characteristic of RPM and thus distills the representative information of the relations existing in the inputs.

On the whole, the SRAN can be formulated in its simplest form as follows:
\begin{equation}
	\begin{aligned}
		\mathbf{r}_{ij}^{(3)} =& \text{SRAN}(\mathbf{M}_i, \mathbf{M}_j) \\
		=& \mathbb{G}(\mathbf{x}_{i}, \mathbf{x}_{j},\mathbf{y}_{i}, \mathbf{y}_{j},\mathbf{z}_{ij}),
	\end{aligned}
\end{equation}
where $\mathbf{r}_{ij}^{(3)}$ is the shared rule embedding of the $\mathbf{M}_i$ and $\mathbf{M}_j$. 
An illustration of SRAN is shown in Figure~\ref{network-SRAN}.

\subsection{Learning and Inference}
With SRAN framework, the question turns to how we train the network, and apply it to infer the correct answer to RPM test. The key to address the question lies in the similarity measure between two rule embeddings, based on which we can define the loss function for SRAN training, and meanwhile determine the best choice during inference.% 

\subsubsection{Similarity function} We introduce similarity function $\mathcal{D}$ to measure the closeness between two rules in the embedding space. 
In this paper, we adopt inner product similarity for good experimental results:
\begin{equation}
	\mathcal{D}(\mathbf{r}, \mathbf{r'})= \mathbf{r}^{\text{T}}\mathbf{r'} .
\end{equation}

\subsubsection{Training}
For a given RPM question, the first two rows $\mathbf{M}_1, \mathbf{M}_2$ are fed into our proposed SRAN and produce the shared rule embedding ${\mathbf{g}}$:
\begin{equation}
	\mathbf{g}=\mathbf{r}_{12}^{(3)} = \text{SRAN}(\mathbf{M}_1, \mathbf{M}_2),
\end{equation}
which represents the dominant pattern of the matrix.

\begin{figure}[tp!]
	\centering
	\includegraphics[width=1.0\linewidth]{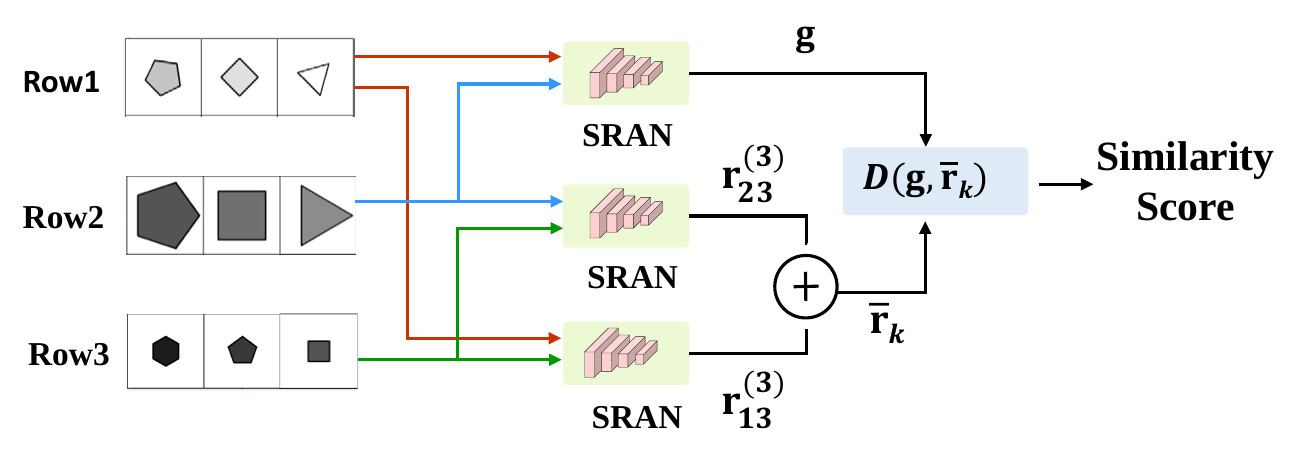}
	\caption{The similarity score for a candidate answer. A multiple-choice panel from the answer set is infilled in the blank panel (row 3), generating a rule embedding $\overline{\mathbf{r}}_k$ through SRAN. The similarity score for the candidate answer can be estimated based on $\overline{\mathbf{r}}_k$ and the dominant rule embedding $\mathbf{g}$ extracted from row 1 and 2}
	\label{network-structure}
	%\vspace{-0.20in}
\end{figure}

Intuitively, the rule extracted from the first two rows can be treated as the reference rule, and we name it the dominant rule in the matrix. Subsequently, the correct answer can be found by checking whether its corresponding rule embedding is similar to the dominant rule. Specifically, given a multiple-choice panel $\omega_k \in \Omega$, where $k \in \{1,..., N\}$, we denote $\overline{\mathbf{r}}_k$ as the new rule embedding inside $\mathbf{M}$ caused by the $k$-th multiple-choice panel:
\begin{equation}
	\overline{\mathbf{r}}_k = \frac{1}{2}\left(\mathbf{r}_{13}^{(3)}+\mathbf{r}_{23}^{(3)}\right).
\end{equation}
This procedure is illustrated in Figure~\ref{network-structure}. In practice, we generate the column-wise rule representation just as the row-wise one, and concatenate the two representations together as the final representation.

For the rule embedding $\overline{\mathbf{r}}^{*}$ generated by rows/columns infilled with correct answer, the desirable SRAN should enforce it to be more similar to the dominant rule ${\mathbf{g}}$, compared to the other rules $\overline{\mathbf{r}}_k$ corresponding to the wrong answers, where $\overline{\mathbf{r}}_k\not=\overline{\mathbf{r}}^*$. Subsequently, the generated rules of $N$ candidates, alongside with the dominant rule, form a tuple containing $N$+1 elements. Based on the similarity function, the ($N$+1)-tuplet loss \cite{sohn2016improved} can be defined for SRAN training:

\begin{equation}
	\mathcal{L} = \text{log}(1+\sum_{k=1, \overline{\mathbf{r}}_k\not=\overline{\mathbf{r}}^*}^{N}\text{exp}(\mathcal{D}({\mathbf{g}}, \overline{\mathbf{r}}_k) - \mathcal{D}({\mathbf{g}}, \overline{\mathbf{r}}^{*}))),
\end{equation}
which means the SRAN can be trained in a fully end-to-end manner.
The architecture of the SRAN (Figure \ref{network-SRAN}) is well matched 
to the problem of abstract reasoning, because it leverages human strategies and explicitly generates the rules governing the matrix.

\subsubsection{Inference}
Once the training of SRAN is finished, we could make the inference of the newly given RPM question. Initially, the intact rows/columns of the RPM are fed into the framework to get the dominant rule $\mathbf{g}$. After that, each multiple-choice panel is filled to the blank position to complete the matrix, and the framework will generate the rule embeddings $\overline{\mathbf{r}}_k$ for all candidate answers, given the current completed matrix. We can accomplish the abstract reasoning by choosing the correct multiple-choice as follows:
\begin{equation}
	k^{*}=\mathop{\arg\max}_{k} \mathcal{D}(\mathbf{g}, \overline{\mathbf{r}}_k).
\end{equation}
Note that since we investigate each panel independently, the above inference process promises that our model's output stays invariant if the answer set is shuffled.

\section{Attribute Bisection Tree for Impartial Dataset}\label{sec:dataset}

RAVEN~\cite{ZhangRAVEN} is a popular RPM-style dataset adopted by all recent studies~\cite{zhanglearning,zheng2019abstract,wang2020abstract}. However, we find severe defects in its answer sets, making RAVEN incompetence as a measurement of abstract reasoning.
In this section, we first give a brief review of RAVEN, and then explain the defects with analysis and experiments.
Finally we introduce a general algorithm to generate an impartial answer set for any attribute-based RPM question. Thus we fix the defects of RAVEN and propose an improved dataset.

\subsection{Defects of RAVEN}\label{exp:dataset}
RAVEN dataset consists of 70,000 RPM questions distributed in 7 different figure configurations.
Panels are constructed with 5 attributes (\verb|Number|, \verb|Position|, \verb|Type|, \verb|Size|, \verb|Color|). Each attribute is governed by one of 4 rules and takes a value from a predefined set. Rules are applied only row-wise in RAVEN.

\begin{figure}[tp!]
	\center
	\includegraphics[width=1.0\linewidth]{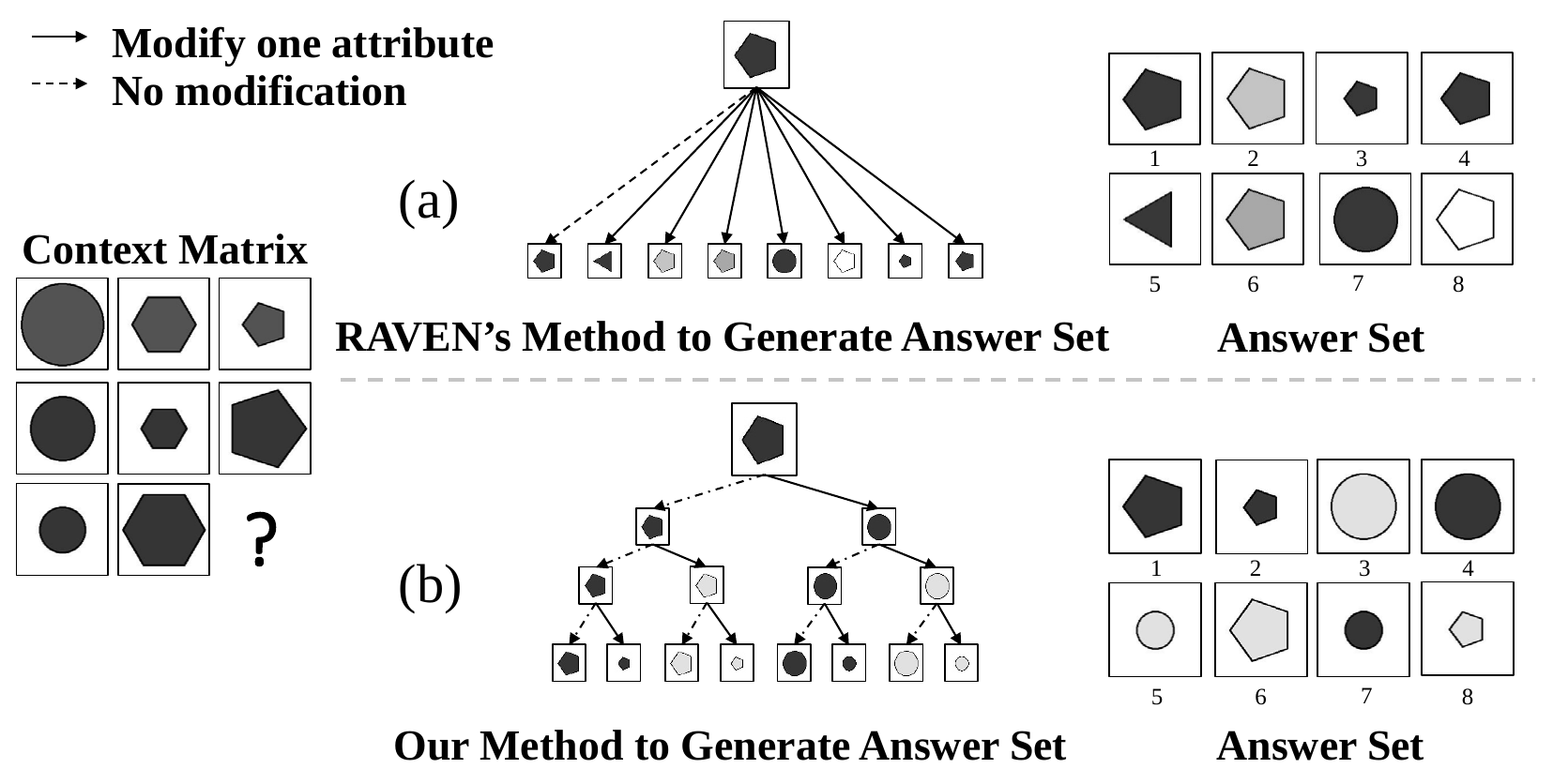}
	\caption{Comparison between RAVEN and I-RAVEN}
	\label{ASR-sample}
	%\vspace{-0.20in}
\end{figure}

After carefully examining the data in RAVEN, we find unexpected pattern among the eight multiple-choice panels. Each distractor in the answer set is generated by randomly modifying one attribute of the correct answer (see Figure \ref{ASR-sample}(a)). 
As a consequence, the panel with the most common values for each attribute will be the correct answer. This means the correct answer can be found by simply scanning the answer set without considering the context images. 
An example is also shown on the right of 
Figure~\ref{ASR-sample}(a). Among the answer set, the most common \verb|Color| and \verb|Type| are black (1, 3, 4, 5, and 7) and pentagon (1, 2, 3, 4, 6, and 8). Besides, multiple-choice panel 1, 2, 5, 6, 7, and 8 are in the same \verb|Size|. Therefore, multiple-choice panel 1, which is the panel with the most common attribute values, is inferred as (and indeed is) the correct answer, even without considering the  context matrix. 

Note that understanding of the context matrix is the cornerstone of RPM test. The RAVEN dataset, where the correct answer can be found without the context, is obviously against the essence of abstract reasoning, and thus is incapable of evaluating abstract reasoning ability.

\begin{table}
	\centering
	\begin{tabular}{lcc} 
		\hline
		{Model}  & { RAVEN} & { I-RAVEN}\\
		\hline
		ResNet~\cite{ZhangRAVEN} & 53.4 & 40.3 \\
		CoPINet~\cite{zhanglearning} & 91.4 & 46.1 \\
		\hline
		Context-blind ResNet  & 71.9 & 12.2 \\ 
		Context-blind CoPINet & 94.2 & 14.2 \\
		\hline
	\end{tabular}
	\caption{Test on RAVEN and I-RAVEN}
	\label{ASR-result}
\end{table}

More severely, such underlying patterns can also be captured by neural networks, especially for models which combine features of eight multiple-choice panels.
We train two models with context-blind~\cite{BarrettMeasuring} setting, including a simple ResNet-based classifier~\cite{ZhangRAVEN} and the competitve CoPINet~\cite{zhanglearning}. These context-blind models are trained with only eight multiple-choice panels as input, and should have predicted the answer randomly, if the dataset is logical.
However, as shown in Table~\ref{ASR-result}, the context-blind models can achieve even better performance to the normal models, which proves that RAVEN contains illogical patterns where the correct answer can be found when only presented with the answer set. 
This type of back-door solutions is quite hidden and has also been discussed in other relational reasoning benchmarks, such as  Visual Question Answering (VQA)~\cite{johnson2017clevr} and symbolic analogy~\cite{HillLearning}.
We can conclude that the 'overfitting' phenomenon on RAVEN reported by~\cite{wang2020abstract} is not caused by a permutation-invariant structure but by the dataset itself.

\begin{algorithm}[h] 
	\renewcommand{\algorithmicrequire}{\textbf{Input:}}
	\renewcommand{\algorithmicensure}{\textbf{Output:}}
	\caption{Attribute Bisection Tree}%算法标题
	\label{alg:dataset}
	\begin{algorithmic}[1]%一行一个标行号
		\REQUIRE the correct answer $\omega^*$
		\STATE Initialize the answer set $\Omega = \{\omega^*\}$
		\STATE Sample $3$ attributes $a_1, a_2, a_3$ according to $\omega^*$
		\STATE Sample new value $v_i$ for each $a_i$
		\FOR{$i=1$ to $3$}
		\STATE Initialize $\Gamma = \{\}$
		\FOR{each $w_k$ in the current answer set $\Omega$}
		\STATE $\gamma \leftarrow$ modifying attribute $a_i$ of $\omega_k$ with $v_i$
		\STATE $\Gamma \leftarrow \Gamma \bigcup \{\gamma\}$
		\ENDFOR
		\STATE $\Omega \leftarrow \Omega \bigcup \Gamma$
		\ENDFOR
		\ENSURE the answer set $\Omega$ ($|\Omega| = 2^3 = 8$)
	\end{algorithmic}
\end{algorithm} 

\subsection{Attribute Bisection Tree}

We design a general algorithm named Attribute Bisection Tree (ABT) to generate an impartial answer set for any attribute-based RPM question.
The ABT ensures attribute modifications among the answer set are well balanced. Thus, no clue can be found to guess the correct answer only depending on the answer set, and no distractor can be eliminated without reasoning from the context matrix as well.

Figure~\ref{ASR-sample}(b) demonstrates the generation process using a tree structure. Each node indicates a multiple-choice panel, and the root of the tree structure is the correct answer. Different levels indicate different iterations, where nodes of this level are the candidate answers of current answer set. The generation process flows in a top-down manner. For each iteration, only one attribute will be modified. At each level, a node has two children nodes, where one node remains the same with the father node, the other changes the value of the attribute sampled for this iteration of the father node. Finally, at the bottom level, we could obtain the whole answer set. Algorithm~\ref{alg:dataset} summarizes the key steps of the answer generation process.

\subsection{I-RAVEN}

\begin{table*}[bhtp]
	\begin{center}
		\begin{tabular}{lc ccccccc} %p{0.85cm}<{\centering}
			\hline
			{Model}  &  
			%\cline{3-10} 
			{ Acc} & {   Center} & { 2$\times$2G} & { 3$\times$3G} & { O-IC} & { O-IG} & { L-R} & { U-D} \\ 
			\hline
			LSTM~\cite{ZhangRAVEN} &  18.9 & 26.2 & 16.7 & 15.1 & 21.9 & 21.1 & 14.6 & 16.5 \\
			WReN~\cite{BarrettMeasuring} & 23.8 & 29.4 & 26.8 & 23.5 & 22.5 & 21.5 & 21.9 & 21.4 \\	
			%\hline
			ResNet~\cite{ZhangRAVEN} &  40.3 & 44.7 & 29.3 & 27.9 & 46.2 & 35.8 & 51.2 & 47.4 \\
			%\hline
			ResNet+DRT~\cite{ZhangRAVEN} &  40.4 & 46.5 & 28.8 & 27.3 & 46.0 & 34.2 & 50.1 & 49.8 \\
			LEN~\cite{zheng2019abstract} & 41.4 & 56.4 & 31.7 & 29.7 & 52.1 & 31.7 & 44.2 & 44.2 \\
			%\hline
			Wild ResNet~\cite{BarrettMeasuring} &  44.3 & 50.9 & 33.1 & 30.8 & 50.9 & 38.7 & 53.1 & 52.6 \\
			%\hline

			CoPINet~\cite{zhanglearning} & 46.1 & 54.4 & 36.8 & 31.9 & 52.2 & 42.8 & 51.9 & 52.5\\
			
			SRAN (Ours) & \textbf{60.8} & \textbf{78.2} & \textbf{50.1} & \textbf{42.4} & \textbf{68.2} & \textbf{46.3} & \textbf{70.1} & \textbf{70.3} \\		
			%\hline				
			\hline
		\end{tabular}
	\caption{Test accuracy of different models on I-RAVEN. Acc denotes the mean accuracy, while other columns show accuracy across seven figure configurations. 2$\times$2G, 3$\times$3G, O-IC, O-IG, L-R, and U-D denote \texttt{2x2Grid}, \texttt{3x3Grid}, \texttt{Out-InCenter}, \texttt{Out-InGrid}, \texttt{Left-Right}, and \texttt{Up-Down}, respectively}
	\label{result-table}
	\end{center}
	%\vspace{-0.2in}
\end{table*}

\begin{table*}[htp]
	%\vspace{-0.2in}
	%\setlength{\tabcolsep}{3pt}
	\begin{center}
		%\vspace{-0.2in}
		\begin{tabular}{lcccccccc} %p{0.85cm}<{\centering}
			\hline
			{Model}  &{ LSTM} &  
			%\cline{3-10} 
			{ ResNet} & { Wild ResNet} & { CoPINet} & { WReN} & { MXGNet} & { LEN} & { SRAN} \\ 
			\hline
			Acc & 35.8 & 42.0 & 48.0 & 56.4 & 62.6 & 66.7 & 68.1  & \textbf{71.3} \\
			\hline
		\end{tabular}
	\caption{Test accuracy of different models on PGM}
	\label{result-table-PGM}
	\end{center}
	%\vspace{-1.0cm}
\end{table*}

With ABT, we generate an alternative answer set for each RPM question in the RAVEN dataset, forming an improved dataset named Impartial-RAVEN (I-RAVEN). Next, we will show that compared with RAVEN, I-RAVEN is more rigorous and fair for evaluating abstract reasoning capability.

Taking Figure~\ref{ASR-sample}(b) for example, each attribute has two different values which distribute evenly in the answer set. 
The attribute \verb|Color| of half answer candidates (1, 2, 4, and 7) are black, while the other half (3, 5, 6, and 8) are light grey. 
Similarly, the attribute \verb|Type| of half answer candidates (1, 2, 6, and 8) are pentagon, while the other half (3, 4, 5, and 7) are circle. 
Half of the answer set (1, 3, 4, and 6) are in the same size, which are different from the other same-sized half (2, 5, 7, and 8).  
As a result, there is no candidate with the most common values for each attribute. In other words, the back-door solution on RAVEN can no longer be appied to the new answer set.

\begin{figure}[t]
    \vspace{0.4in}
	\center
	\includegraphics[width=1.0\linewidth]{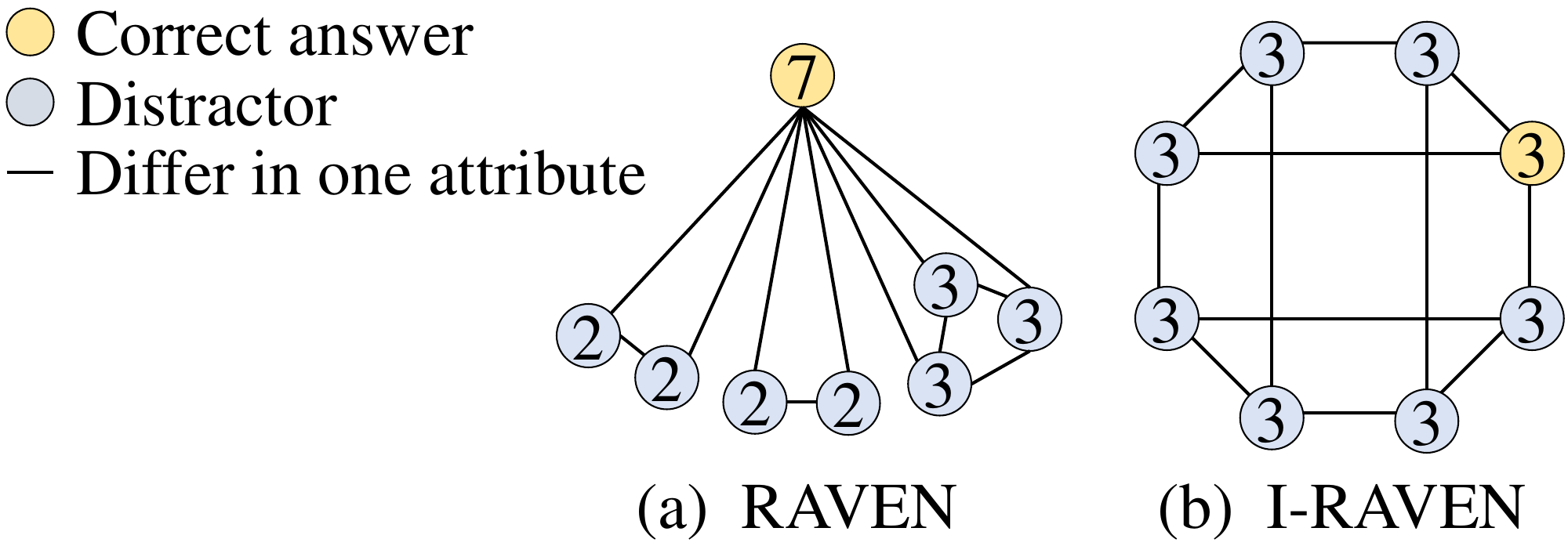}
	\caption{Characterizing answer sets of RAVEN and I-RAVEN using graphs}
	\label{topology}
	%\vspace{-0.25in}
\end{figure}

To better explain the superiority of I-RAVEN over RAVEN, as shown in Figure~\ref{topology}, we use undirected graphs to characterize typical answer sets of the two datasets respectively, where each candidate answer is represented by a node with its degree infilled. An edge between two nodes represents that the corresponding candidates differ in one attribute.
In Figure~\ref{topology}(a), there is always a central node with a degree of 7 and the other nodes with less degrees.
The back-door solution is to find the central node, which is indeed the correct answer.
By contrast, in Figure~\ref{topology}(b), due to balanced  modifications of attributes, each node always has the same degree of 3, which is indistinguishable from each other without the context matrix.
Moreover, inspired by~\cite{HillLearning}, we make the noise attribute \texttt{Uniformity} of each distractor stay consistent with the correct answer, so that each distractor is perceptually plausible and cannot be eliminated simply by attribute mismatching. This setting encourages models to reason from the context.

We also train ResNet and CoPINet on I-RAVEN alongside with their context-blind versions to verify the fairness of the proposed dataset.
The right column in Table~\ref{ASR-result} lists the results  which are in stark contrast with those on the original RAVEN dataset.
The performance of context-blind models is almost at a random guess level (12.5\%), while the normal models relying on the context can perform much better.

\section{Experiments}

\subsection{Experimental Setup}\label{experimental setup}
\label{models}

With I-RAVEN, we first compare our method with several state-of-the-art models using public implementations, including LSTM~\cite{hochreiter1997long}, ResNet-based~\cite{he2016deep} image classifier (ResNet), ResNet with DRT~\cite{ZhangRAVEN}, Wild ResNet~\cite{BarrettMeasuring}, WReN~\cite{BarrettMeasuring}, CoPINet~\cite{zhanglearning}, and LEN~\cite{zheng2019abstract}.
We adopt the public implementations of LSTM, ResNet, and DRT in~\cite{ZhangRAVEN}. Other variants of LEN are not included in this section because they require additional training labels.

PGM~\cite{BarrettMeasuring} is another RPM dataset consisting of 1.42M questions. Rules in a matrix are composed with 1 to 4 relation-object-attribute tuples and can be applied along the rows or columns. 
SRAN is compared with results on PGM reported in~\cite{BarrettMeasuring,zhanglearning,zheng2019abstract,wang2020abstract}.

For our SRAN, we adopt three ResNet-18~\cite{he2016deep} as the embedding networks for the three hierarchies, by modifying the input channels. The gate fusion $\varphi_1$ and $\varphi_2$ are 2-layer fully connected networks, while $\varphi_3$ is a 4-layer fully connected network with dropout~\cite{srivastava2014dropout} of 0.5 applied on the last layer. 
We adopt stochastic gradient descent using ADAM~\cite{kingma2014adam} optimizer. The exponential decay rate parameters are $\beta_1=0.9$, $\beta_2=0.999$, $\epsilon=10^{-8}$.
Each reported accuracy is averaged over 5 runs.

\subsection{Comparisons with State-of-the-art Methods}\label{sec:comparisons}

Table~\ref{result-table} and Table~\ref{result-table-PGM} list the test accuracy of different models trained on I-RAVEN and PGM, respectively. From the table, it is obvious that our proposed SRAN outperforms other methods by a considerable margin. 
Besides, we observe that models benefit from inductive biases, such as the competitive CoPINet, LEN, Wild ResNet, and our SRAN. Such inductive biases (even partly) can encourage models to explore the underlying rules. 
For more detailed comparison, Table~\ref{result-table} also reports the accuracy on seven figure configurations of I-RAVEN. We can observe that accuracy on different configurations is not uniform, possibly due to the difficulty of configurations. But compared with other models, our SRAN consistently achieves the best performance on all the configurations, which proves that our model can work stably, even facing diverse conditions and complex rules. 

We observe that the accuracy of SRAN on I-RAVEN is very close to that on the original RAVEN dataset (60.8\% vs. 60.7\%). This phenomenon is as expected since our method mainly focuses on rules in the context, and thus is robust to the answer set.

\subsection{Ablation Study}
As aforementioned, our method mainly gains from the inductive-biased architecture. To validate this point, we study the effects of different components in our SRAN. Table \ref{ablation-table} lists the results.

We analyze the stratified strategy to incrementally induce rules, using the performance of different choices of hierarchies.
Specifically, we set the rule embedding of certain hierarchy as a zero vector before gate function $\varphi$. Thus, the gate function regulates the flow of features into the gated embedding fusion module.
We observe that combing more hierarchies always leads to better performance on I-RAVEN, which shows all hierarchies contribute to our framework.
We make $\mathbb{E}_{\text{cell}}$ orderless by summing all cell-wise embeddings and observe a major drop in performance.
These observations prove the effectiveness of the proposed inductive biases of order sensitivity and incremental rule induction.

\begin{figure}[htp!]
	\centering
	\includegraphics[width=1.0\linewidth]{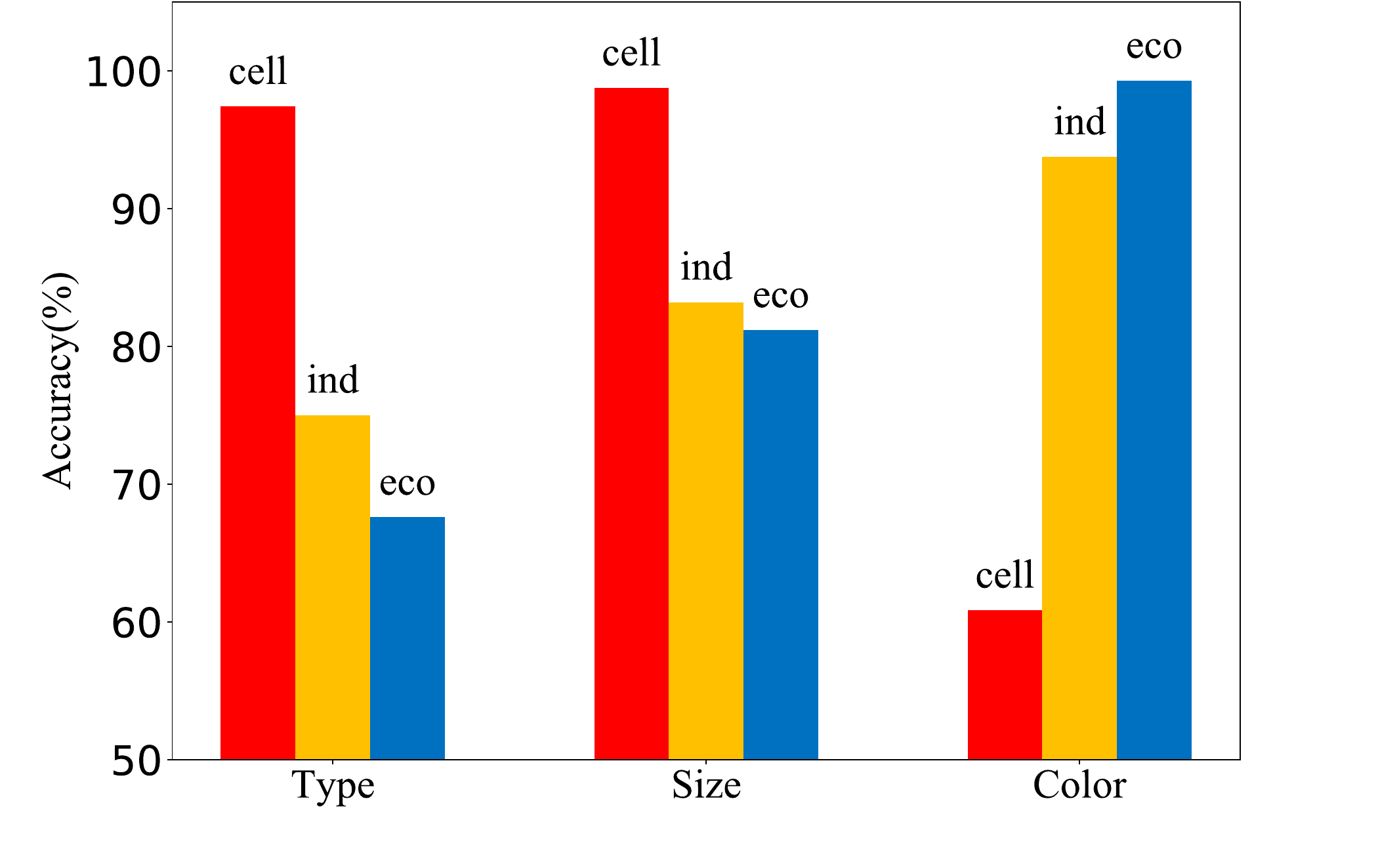}
	%\vspace{-0.5cm}
	\caption{Accuracy of single hierarchy with respect to the different attributes}
	\label{single-level-attribute}
\end{figure}

We further discover that the effectiveness of our model can be put down to its attention to different attributes.
We conduct experiments only utilizing single-hierarchy rule embeddings from $\mathbb{E}_{\text{cell}}, \mathbb{E}_{\text{ind}}, \mathbb{E}_{\text{eco}}$, with respect to three attributes (\verb|Type|, \verb|Size|, and \verb|Color|) of I-RAVEN. As shown in Figure~\ref{single-level-attribute}, $\mathbb{E}_{\text{cell}}$ has strong capacity to infer attributes \verb|Type| and \verb|Size|, but struggles to distinguish attribute \verb|Color|. By contrast, $\mathbb{E}_{\text{ind}}$ and $\mathbb{E}_{\text{eco}}$ have modest ability to infer attributes \verb|Type| and \verb|Size|, and are efficient for attribute \verb|Color|.

\subsection{The Advantage of Rule Embeddings}\label{sec:interpretability}

In the real RPM test,  it is not clear whether the rule exists in rows or columns. Therefore, it is important to check whether the proposed model can discover the knowledge without any guidance. 
Rule induction for columns is normally left out when trained on I-RAVEN, given the prior knowledge that rules are applied only row-wise.
In order to test the ability of distinguishing whether the rules are applied along rows or columns, we train a SRAN model on I-RAVEN which the induction for column rules is reintegrated into.
As a result, there is only a bit drop in accuracy (from 60.8\% to 59.6\%). This indicates that our model can neglect the distraction brought by columns on its own.

\begin{table}[htp]
	\centering
	\begin{tabular}{lc} 
		\hline
		{Model}   & { I-RAVEN} \\	
		\hline
		$\mathbb{E}_{\text{cell}}$ (orderless) &  23.5 \\
		$\mathbb{E}_{\text{cell}}$ &  36.7 \\
		$\mathbb{E}_{\text{ind}}$ &  48.7 \\
		$\mathbb{E}_{\text{eco}}$&  51.6 \\
		$\mathbb{E}_{\text{cell}}$ + $\mathbb{E}_{\text{eco}}$ &  52.9 \\
		$\mathbb{E}_{\text{ind}}$ + $\mathbb{E}_{\text{eco}}$ &  57.0 \\
		$\mathbb{E}_{\text{cell}}$ + $\mathbb{E}_{\text{ind}}$&  57.8 \\
		$\mathbb{E}_{\text{cell}}$ + $\mathbb{E}_{\text{ind}}$ + $\mathbb{E}_{\text{eco}}$   & \textbf{60.8} \\	
		\hline
	\end{tabular}
	\caption{SRAN ($\mathbb{E}_{\text{cell}}$+$\mathbb{E}_{\text{ind}}$+$\mathbb{E}_{\text{eco}}$) and the results of eliminating different components}
	\label{ablation-table}
\end{table}

\section{Conclusion}
In this paper, we introduced necessary inductive biases for abstract visual reasoning task, such as order sensitivity and incremental rule induction. We further proposed a novel Stratified Rule-Aware Network, which could extract multiple granularity rule embeddings at different level and integrate them through a gated embedding fusion module. A rule similarity metric was further introduced based on the embeddings, so that SRAN can not only be trained using a tuplet loss but also infer the best answer according to the similarity score.
We also designed an algorithm named Attribute Bisection Tree to fix the defects of the popular dataset RAVEN, and generated a more rigorous dataset based on the algorithm. Extensive experiments conducted on PGM dataset and our improved dataset I-RAVEN proved that, our proposed framework could significantly outperform other state-of-the-art approaches. Moreover, we studied the effects of each component of our proposed model and evaluated the advantage of our induced rule embeddings.

\section{Acknowledgments}

This work was supported by National Natural Science Foundation of China (62022009, 61872021), Beijing Nova Program of Science and Technology (Z191100001119050), State Key Lab of Software Development Environment (SKLSDE-2020ZX-06), Fundamental Research Funds for Central Universities (YWF-20-BJ-J-646), and the Academic Excellence Foundation of BUAA for PhD Students.

\bibliography{egbib}
\end{document}